\documentclass[letterpaper]{article} 
\usepackage{aaai24}  
\usepackage{times}  
\usepackage{helvet}  
\usepackage{courier}  
\usepackage[hyphens]{url}  
\usepackage{graphicx} 
\usepackage{natbib}  
\usepackage{caption} 
\usepackage{algorithm}
\usepackage{algorithmic}

\usepackage{booktabs} 
\usepackage{epsfig}
\usepackage{amsmath}
\usepackage{amssymb}
\usepackage{colortbl}  
\usepackage{xcolor}
\usepackage{array}
\usepackage{subfigure}

\definecolor{lightred}{RGB}{255,182,203}
\definecolor{lightorange}{RGB}{255,190,102}
\definecolor{lightyellow}{RGB}{255,255,165}

\usepackage{newfloat}
\usepackage{listings}
\DeclareCaptionStyle{ruled}{labelfont=normalfont,labelsep=colon,strut=off} 
\floatstyle{ruled}
\newfloat{listing}{tb}{lst}{}
\floatname{listing}{Listing}
\title{SpectralNeRF: Physically Based Spectral Rendering with Neural Radiance Field}
\author{
    Ru Li\textsuperscript{\rm $1$},
    Jia Liu\textsuperscript{\rm $2$},
    Guanghui Liu\textsuperscript{\rm $2*$},
    Shengping Zhang\textsuperscript{\rm $1$} ,
    Bing Zeng\textsuperscript{\rm $2$},
    Shuaicheng Liu\textsuperscript{\rm $2$}\footnote{Corresponding authors}
}
\affiliations{
    \textsuperscript{\rm 1}Harbin Institute of Technology, Weihai, China\\
    \textsuperscript{\rm 2}University of Electronic Science and Technology of China, Chengdu, China\\

    \{liru, s.zhang\}@hit.edu.cn, \{liujia21@std., guanghuiliu@, eezeng@, liushuaicheng@\}uestc.edu.cn

}

\usepackage{bibentry}

\begin{document}

\maketitle

\begin{abstract}
In this paper, we propose SpectralNeRF, an end-to-end Neural Radiance Field (NeRF)-based architecture for high-quality physically based rendering from a novel spectral perspective. We modify the classical spectral rendering into two main steps, 1) the generation of a series of spectrum maps spanning different wavelengths, 2) the combination of these spectrum maps for the RGB output. Our SpectralNeRF follows these two steps through the proposed multi-layer perceptron (MLP)-based architecture (SpectralMLP) and Spectrum Attention UNet (SAUNet). Given the ray origin and the ray direction, the SpectralMLP constructs the spectral radiance field to obtain spectrum maps of novel views, which are then sent to the SAUNet to produce RGB images of white-light illumination. Applying NeRF to build up the spectral rendering is a more physically-based way from the perspective of ray-tracing. Further, the spectral radiance fields decompose difficult scenes and improve the performance of NeRF-based methods. Comprehensive experimental results demonstrate the proposed SpectralNeRF is superior to recent NeRF-based methods when synthesizing new views on synthetic and real datasets. The codes and datasets are available at {\url{https://github.com/liru0126/SpectralNeRF}}.
\end{abstract}

\section{Introduction}
\label{sec:intro}

Newton found that white light can be dispersed into a series of spectrums with various colors from red to violet by passing light through the glass prism~\cite{newton7serie}.
After that, the research on spectral theory developed rapidly~\cite{pickholtz1982theory,helffer2013spectral}. Until now, the application of spectral images has involved various aspects of daily life, including object detection~\cite{liang2018material}, face recognition~\cite{uzair2015hyperspectral}, and so on. Spectral images can record and reveal the electromagnetic radiation intensity information of objects, which is an important interdisciplinary subject mainly involving physics and chemistry~\cite{bertrand2021deciphering}. 

Spectral rendering is a fundamental problem in computer graphics, which can understand the absorption, reflection, and other interactions with objects and has been used to generate photo-realistic images~\cite{peercy1993linear}. Conventional spectral rendering involves two transformations: 1) the spectral power distribution $L$ to the XYZ image, achieved by using integral operation through the visible light; 2) the XYZ image to the RGB image, realized by the conversion matrix~\cite{smits1999rgb}. 
Based on the transformations of $L$ $\to$ XYZ $\to$ RGB, physically-based spectral rendering has been widely researched over the past decades~\cite{watanabe2013performance,sun2001spectrally,knaus2011progressive}.
Such methods predict photo-realistic rendering that no effect that contributes to the interaction of light with a scene is neglected. However, they merely generate one image of the current viewpoint and are limited in representing the scene.

\begin{figure}[t]
   \centering
   \includegraphics[width=1.0\linewidth]{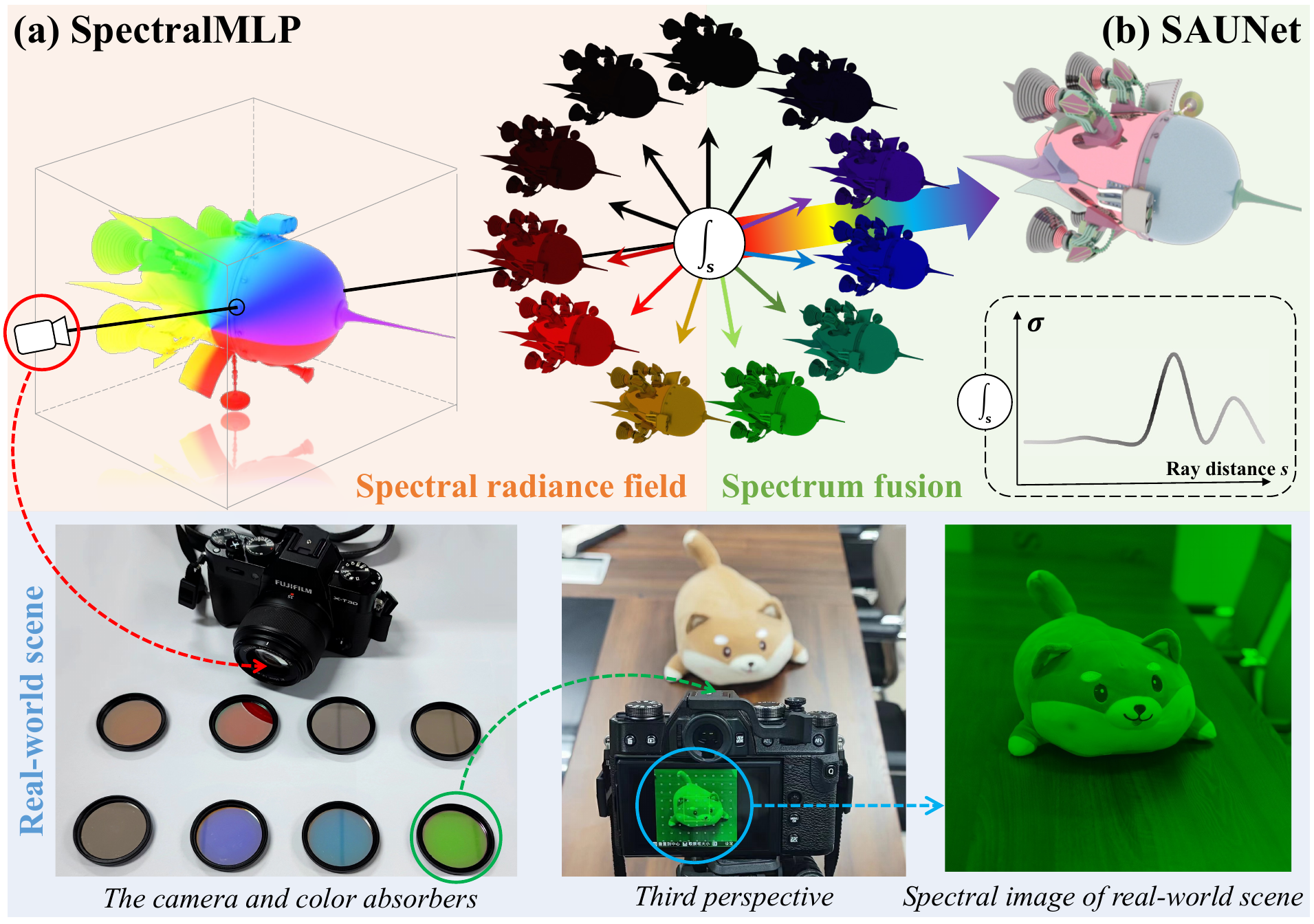}\\
   \caption{SpectralNeRF builds up the process of spectral rendering. The first step samples spectrum points along the ray and uses volume rendering to generate spectrum maps. 
   The second step fuses discrete spectrum maps to obtain RGB images. 
   \textit{Up:} our pipeline. \textit{Bottom:} our capture device.
   }
   \label{fig:intro}
\end{figure}

Recently, Neural Radiance Field (NeRF) is designed to render compelling images of 3D scenes from novel viewpoints~\cite{mildenhall2020nerf}. NeRF-based methods achieve photo-realistic rendering of scenes by encoding the volumetric density and color of a scene within the weights of a coordinate-based multi-layer perceptron (MLP). Subsequently, a series of works focused on recovering the radiance field using deep neural networks~\cite{barron2021mip,barron2022mip,yan2023nerf}.
This approach has enabled significant progress toward photo-realistic view synthesis and can solve the limitation that spectral rendering cannot represent the overall scene. However, NeRF-based methods may lose important details when encountering complicated scenes. 

In this paper, we propose a NeRF-based architecture to achieve the physically-based spectral rendering, named SpectralNeRF. We construct the spectral radiance field for the scene information along the ray, which facilitates the rendering process and keeps the rendering pipeline simple yet effective by employing a physically-based multi-spectral integral calculation for the $L$ $\to$ XYZ $\to$ RGB conversion. For spectral rendering, applying the neural radiance field to learn the spectrum maps and using the integration of spectral bands are more physically-based ways.
For NeRF-based methods, complicated scenes can be simplified through multiple spectral components compared to only normal RGB images, which is superior to previous NeRF-based methods in terms of geometry and texture reconstruction under complex scenes and image quality of the novel viewpoint synthesis. 
The motivation, that is, the need for spectral radiance fields is: spectral information can provide more details on the material constitution of objects in the scene, which have been utilized in classic vision tasks, such as material classification~\cite{jumanazarov2022significance}. 
The idea of importing spectral information to rendering is a new perspective, which may provide inspiration for rendering and vision tasks.

Theoretically, we modify the traditional spectral rendering pipeline into two steps. According to the variant rendering pipeline, we first design an MLP-based architecture, which maps from an input 5D coordinate (3D position and 2D viewing direction) of real and synthetic scenes to properties of the scene (volume density and spectral radiance) at that location, named SpectralMLP. Volume rendering is applied to composite these values into discrete spectrum maps (Fig.~\ref{fig:intro} (a)). Then, SAUNet is proposed to combine these spectrum maps into high-quality RGB images (Fig.~\ref{fig:intro} (b)). In order to extract the spectral information better, we design the Spectrum Attention (SA) module to better explore the correlations between spectrum maps. The pipeline can be transferred to existing NeRF-based methods to promote their performance if we select their architecture as the baseline of SpectralMLP. 
We capture the real-world scenes and render the spectral datasets that contain spectrum maps and RGB images to optimize the outputs of SpectralMLP and the SAUNet, respectively.

Overall, our contributions can be summarized as:
\begin{itemize}

\item We propose SpectralNeRF, that builds up physically-based rendering with NeRF from the spectral perspective, which leads to mutual enhancement of spectral rendering and NeRF-based methods. 

\item We design the SAUNet to fuse the discrete spectrum maps to generate high-quality RGB images, which can approach the integral calculation in spectral rendering.

\item We render 8 spectral datasets and capture 2 real-world scenes with spectrum maps and RGB images,
and provide comprehensive comparisons of these datasets with several NeRF-based methods to demonstrate the superiority of the SpectralNeRF.
\end{itemize}


\section{Related Works}
\textbf{Neural Radiance Field for 3D Scenes.}
Using the neural network to represent a 3D scene and generate novel views with weights of MLP or other network parameters has become a hot topic. Previous methods address the issue with explicit discrete representations~\cite{dai2015random,aliev2020neural,wu2020adversarial,mildenhall2019local,waechter2014let}.
Since Mildenhall~\emph{et al.} introduced the differentiable volumetric rendering technique to optimize a neural radiance field~\cite{mildenhall2020nerf}, a number of studies have been carried out to dive deeper into NeRF-based architecture, including
more detail preservation methods~\cite{barron2021mip,barron2022mip,chen2022aug,dave2022pandora},
the faster training and inference of NeRF~\cite{martin2021nerf,reiser2021kilonerf}, 
the extension from image to video~\cite{li2021neural,li2023dynibar}, 
the refractive novel-view synthesis~\cite{bemana2022eikonal},
the dynamic scenes~\cite{cao2023hexplane}, 
the LiDAR scenes~\cite{huang2023neural},
the infrared and spectral scenes~\cite{poggi2022cross},
and the event cameras~\cite{rudnev2023eventnerf}.
It is challenging for these methods to represent scenes with complex textures. 
To solve such problems, we present a novel NeRF-based architecture, which introduces spectral information into the radiance field to simplify complicated scenes.

\textbf{Spectral Rendering.}
With the development of computing power, various rendering technologies are proposed to obtain photo-realistic images~\cite{nguyen2018rendernet,peters2019using,dai2020pbr,li2022phyir,li2023pbr,hu2023point2pix}.
Spectral rendering is a more physically correct technique that indeed models a scene's light transport with real wavelengths~\cite{wilkie2002tone}. Over the past decades, many physically-based spectral rendering methods have been proposed, including stochastic sampling over the visible light~\cite{watanabe2013performance}, representing spectral information using Fourier coefficients~\cite{peters2019using} and sampling in the spatial domain~\cite{knaus2011progressive,watanabe2013performance}. These methods are designed for canonical ray-tracing rendering pipelines, which might be time-consuming when rendering scenes with complicated geometry. We consider the scene information along the ray and take advantage of NeRF-based architecture to combine the neural radiance field and spectral information to perform physically-based spectral rendering.


\section{Preliminaries of Spectral Rendering}\label{sec:spectral_rendering}

Given the CIE tristimulus values X, Y and Z, the CIE color matching functions $f_X(\lambda)$, $f_Y(\lambda)$ and $f_Z(\lambda)$ involve the influence of light with wavelength $\lambda$ to the three values~\cite{cvrl}, which were defined by measuring the mean color perception of a sample of human observers over the visual range from ${\lambda}_{\text{violet}}=380$ to ${\lambda}_{\text{red}}=780$ nanometer ($nm$). The following equation calculates the CIE X, Y, and Z values for light with wavelength $\lambda$:
\begin{equation}
\small
\label{eq:xyz_discrete}
\left\{\begin{array}{l}
X=\kappa \sum  f_X(\lambda) L(\lambda) \Delta \lambda \\
Y=\kappa \sum  f_Y(\lambda) L(\lambda) \Delta \lambda \\
Z=\kappa \sum  f_Z(\lambda) L(\lambda) \Delta \lambda,
\end{array}\right.
\end{equation}
where $\kappa$ is a normalizing constant, $\sum$ represents the summation of visible light, $\Delta \lambda$ represents the sampling interval, and $L$ denotes the spectral power distribution of the light source from the direction  ($\theta_{\mathrm{v}}, \varphi_{\mathrm{v}}$) of observation $x$:
\begin{equation}
\small
\label{eq:l}
L\left(x, \theta_{\mathrm{v}}, \varphi_{\mathrm{v}}, \lambda\right)\!=\!\int_{\Omega}  f_{\mathrm{r}}\left(x, \theta, \varphi, \theta_{\mathrm{v}}, \varphi_{\mathrm{v}}, \lambda\right) R_{\mathrm{i}}(x, \theta, \varphi, \lambda) \cos \theta \mathrm{~d} \omega,
\end{equation}
where $R_{\mathrm{i}}$ represents the radiance from direction ($\theta, \varphi$) to point $x$, $\Omega$ is the hemispherical space on the surface where point $x$ is located, 
$f_{\mathrm{r}}$ represents the bidirectional reflectance distribution function (BRDF), which is determined by the reflection characteristics of the material at point $x$,  
$\mathrm{~d} \omega$ is a solid angle.
Note that, Eq.~\ref{eq:xyz_discrete} is calculated with the form of summation to estimate the original continuous integral.

\begin{figure}[t]
   \centering
   \includegraphics[width=1.0\linewidth]{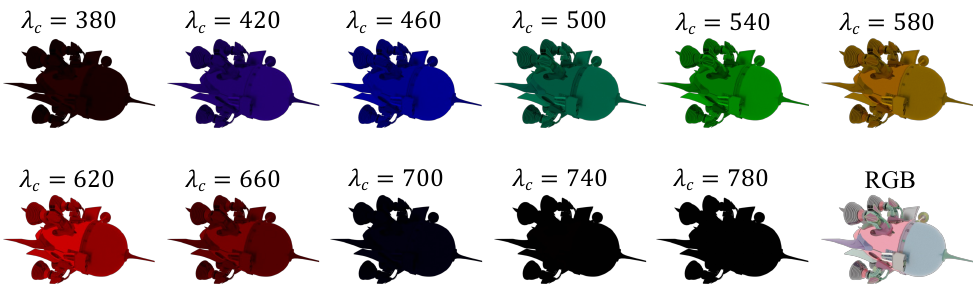}\\
   \caption{
   The RGB spectrum maps of different wavelengths and the RGB image of white-light illumination.}
   \label{fig:dataset_spaceship}
\end{figure}

To obtain a colorimetrically correct RGB image, the X, Y, and Z 
values are transformed to the sRGB color space using:
\begin{equation}
\small
\label{eq:rgb_trans}
\left[\begin{array}{l}
R \\
G \\
B
\end{array}\right]=\left[\begin{array}{ccc}
3.133 & -1.616 & -0.490 \\
-0.978 & 1.916 & 0.033 \\
0.072 & -0.229 & 1.405
\end{array}\right]\left[\begin{array}{l}
X \\
Y \\
Z
\end{array}\right].
\end{equation}

Due to inconsistency of application environments, there are various XYZ $\to$ RGB conversion methods~\cite{smits1999rgb}. The matrix $M^{\text{c}}$ listed here is computationally convenient.

Obtaining $L$ from the rendering machine is difficult. In order to simplify the rendering pipeline of $L$ $\to$ XYZ $\to$ RGB, we embed the Eq.~\ref{eq:rgb_trans} to Eq.~\ref{eq:xyz_discrete} to obtain the RGB spectrum maps corresponding to the wavelengths. Linearly combining the two equations to generate the RGB spectrum maps is reasonable on the one hand, and is simple yet effective on the other hand. The RGB spectrum maps corresponding to the wavelengths can be formulated as:
\begin{equation}
\small
\label{eq:rgb_discrete}
\left\{\begin{array}{l}
R_{\lambda} \! = \! (M_{11}^{\text{c}} f_X(\lambda) \!+ \! M_{12}^{\text{c}}f_Y(\lambda) \! + \! M_{13}^{\text{c}} f_Z(\lambda)) L(\lambda) \Delta \lambda \\
G_{\lambda} \! = \! (M_{21}^{\text{c}} f_X(\lambda) \! + \! M_{22}^{\text{c}}f_Y(\lambda) \! + \! M_{23}^{\text{c}} f_Z(\lambda)) L(\lambda) \Delta \lambda \\
B_{\lambda} \! = \! (M_{31}^{\text{c}} f_X(\lambda) \! + \! M_{32}^{\text{c}}f_Y(\lambda) \! + \! M_{33}^{\text{c}} f_Z(\lambda)) L(\lambda) \Delta \lambda.
\end{array}\right.
\end{equation}

Finally, the RGB spectrum maps are combined to generate the RGB image of white-light illumination:
\begin{equation}
\small
\label{eq:rgb}
\left\{\begin{array}{l}
R= \kappa \sum R_{\lambda}  \\
G= \kappa \sum G_{\lambda}  \\
B= \kappa \sum B_{\lambda}.
\end{array}\right.
\end{equation}

Figure~\ref{fig:dataset_spaceship} shows an example that includes 11 RGB spectrum maps and
one RGB image rendered with our hypothesis by Mitsuba~\cite{mitsuba}. The $\lambda_c$ in Fig.~\ref{fig:dataset_spaceship} represents the center of the sampling interval of spectral illuminates. 

\begin{figure*}[t]
   \centering
   \includegraphics[width=0.99\linewidth]{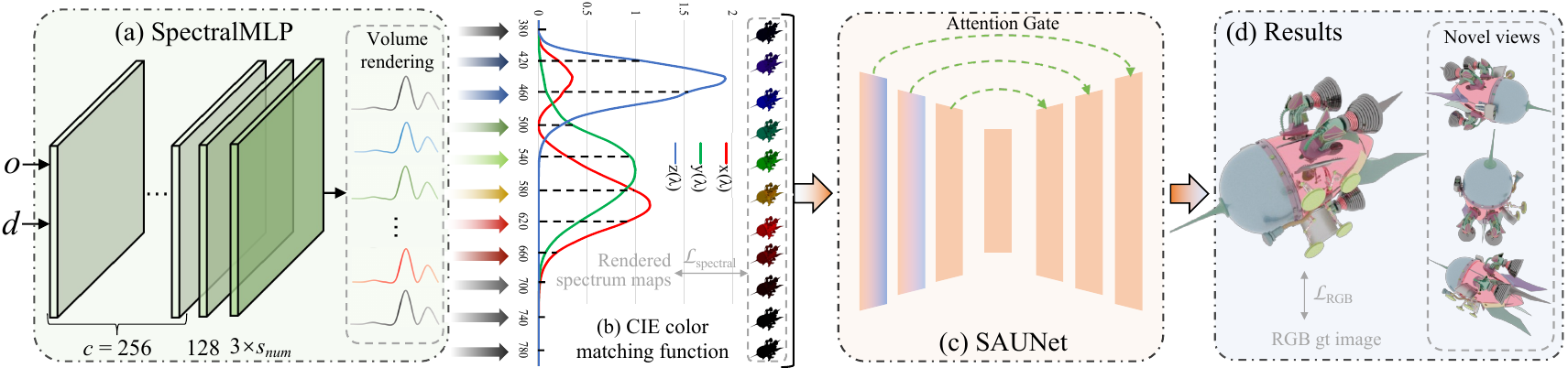}\\
   \caption{An overview of the SpectralNeRF. We first design (a) SpectralMLP to construct the spectral radiance field and generate $s_\text{num}$ RGB spectrum maps with novel views using volume rendering. The generated spectrum maps are constrained by the rendered spectral images. The color of the spectrum maps matches the distribution of the CIE color matching function in (b). The (c) SAUNet combines the discrete spectrum maps to produce high-quality RGB outputs, constrained by the rendered RGB images. $\mathbf{o}$ is the ray origin, $\mathbf{d}$ is the ray direction and $c$ represents the channels of different layers in SpectralMLP.
}
   \label{fig:pipeline}
\end{figure*}


\section{Method}

We propose an end-to-end NeRF-based architecture to achieve the physically-based spectral rendering from a novel perspective. As shown in Fig.~\ref{fig:pipeline}, the architecture includes two modules operating the RGB spectrum map rendering (Fig.~\ref{fig:pipeline} (a)) and the spectrum fusion (Fig.~\ref{fig:pipeline} (c)). The first module is an MLP-based network to produce spectrum maps according to the given ray origin $\mathbf{o}$ and ray direction $\mathbf{d}$. The second module fuses the discrete spectrum maps to obtain an RGB image of white-light illumination, which applies the attention mechanism to better extract useful information from spectrum maps to approach the integral calculation in spectral rendering. Note that, we generate $s_\text{num}$ discrete spectrum maps by uniform sampling through the visible light 380$nm$--780$nm$. The overall color of the generated spectrum maps conforms to the distribution of the CIE color matching function in Fig.~\ref{fig:pipeline} (b). Both spectral rendering and NeRF-based methods will be improved through the proposed pipeline.

\subsection{Spectral Radiance Field}

We represent the scene as spectral radiance fields within bounded 3D volumes. For a given ray origin $\mathbf{o}=(x,y,z)$ and ray direction $\mathbf{d}=(\theta, \phi)$, we propose the SpectralMLP $F_{\boldsymbol{\Theta}}$ to generate the spectral radiance $s_{\lambda_i}$ and the density $\sigma$ of the ray $\mathbf{r}(t)=\mathbf{o}+t\mathbf{d}$. SpectralMLP $F_{\boldsymbol{\Theta}}$ achieves the mapping from $(\mathbf{o},\mathbf{d})$ to ($s_{\lambda_i},\sigma$), which is defined as: 
\begin{equation}
\small
(\mathbf{s}_{\lambda_i}, \sigma)=F_{\boldsymbol{\Theta}}\left(\gamma(\mathbf{o}), \gamma(\mathbf{d})\right),
\end{equation}
where $i\in \left \{ 1,2,...,s_{\text{num}} \right \}$, 
$\gamma$ represents the positional encoding~\cite{rahaman2019spectral} that maps the inputs into higher dimensional frequency space, which is applied separately to each of the three coordinate values in $\mathbf{x}$ and to the three components of the direction unit vector $\mathbf{d}$. The $s_{\text{num}}$ is set to 11 to achieve the balance between performance and efficiency.

The SpectralMLP finally outputs $s_{\text{num}}$ spectral radiance and one $\sigma$ value. As for different wavelength $\lambda$, the density of each point is same, but the spectral radiance is different. Volume rendering~\cite{levoy1990efficient} is then 
applied to render the spectral radiance $s_{\lambda_i}$ of each ray passing through the scene. The spectrum value $\widehat{\mathbf{S}}_{\lambda_i}(\mathbf{r})$ of ray $\mathbf{r}(t)$ is computed as:
\begin{equation}
\small
\widehat{\mathbf{S}}_{\lambda_i}(\mathbf{r})=\int_{t_n}^{t_f} T(t) \sigma(\mathbf{r}(t)) \mathbf{s}_{\lambda_i}(\mathbf{r}(t), \mathbf{d})dt,
\end{equation}
where $T(t)=\exp \left(-\int_{t_n}^t \sigma(\mathbf{r}(p)) d p\right)$, $t$ denotes a position along the ray, $t_n$ and $t_f$ are the near and far boundary.

\begin{figure}[t]
   \centering
   \includegraphics[width=0.95\linewidth]{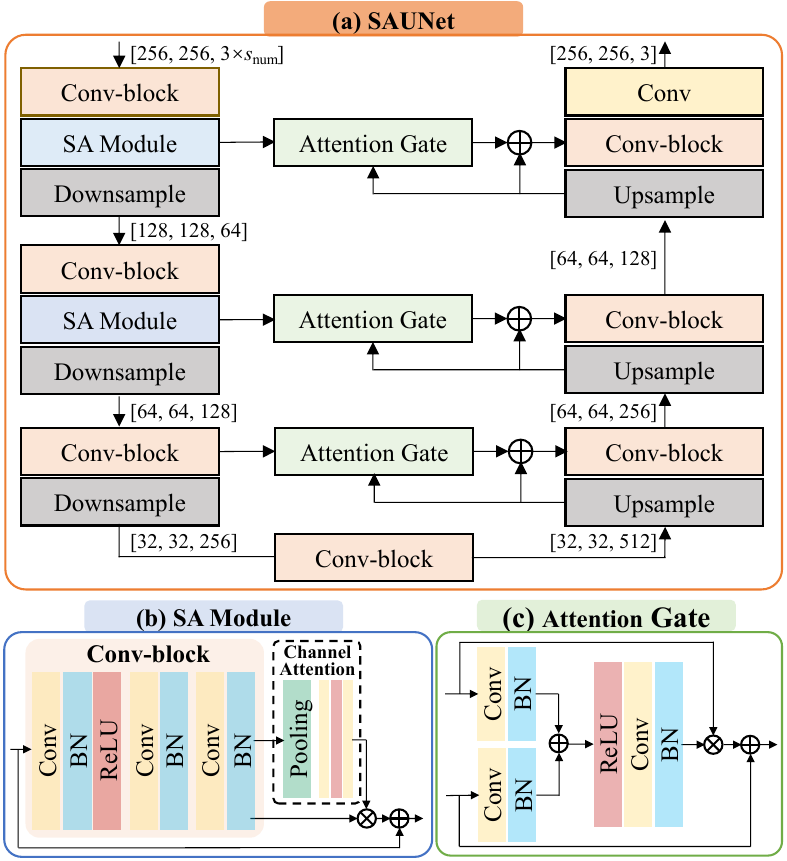}\\
   \caption{The detailed architecture of SAUNet.
   }
   \label{fig:saunet}
\end{figure}

\subsection{SAUNet}
\label{sec:saunet}
Theoretically, according to Eq.~\ref{eq:rgb}, directly combining the spectrum maps of every wavelength among the visible light is practicable. Rendering each spectral dataset with around 400 images 
may be adequate for the linear combination. However, in our implementation, such an operation is insufficient because the spectral datasets are extremely sparse. 
Therefore, we propose the Spectrum Attention UNet (SAUNet) (Fig.~\ref{fig:saunet} (a)) to learn the correlations of spectrum maps and generate high-quality RGB outputs. 
Applying the SAUNet can imitate the original integral operation better and produce results close to the ground truth. We first introduce the Attention Gate~\cite{oktay2018attention} (Fig.~\ref{fig:saunet} (c)) to refine features from the encoder. The high-level features before outputting to the next decoder stage are guided by the low-level features from the encoder using the attention mechanism.
We further design a Spectrum Attention (SA) module to better explore the correlations of spectrum maps. 

The standard residual convolutional network is insufficient for extracting the spectral dependencies~\cite{jiang2022less}. We modify the standard residual blocks and introduce the SA module (Fig.~\ref{fig:saunet} (b)) to better combine the discrete spectrum maps to make the results close to the integral calculation in spectral rendering.
Specifically, three $1 \times 1$ convolutional blocks are first used to reorganize and reweight the importance of spectrum maps. The channel attention (CA)~\cite{hu2018squeeze} is then introduced to focus on inter-spectral feature fusion by attention mechanism in the channel dimension.

SAUNet contains 3 encoders and 3 decoders. Skip connections with Attention Gate pass feature maps from each encoder to decoder. Feature maps from low-levels contain more detailed information of spectrum maps. The SA module is placed in the first two encoders to best use the spectral information because higher-level features are more abstract, which may affect the ability of the network to explore the correlations. 
The fusion process is defined as:
\begin{equation}
\small
\widehat{C}=\text{SAUNet}(\widehat{\mathbf{S}}_{\lambda_i}),
\end{equation}
where $\widehat{C}$ represents the final output RGB image.

\subsection{Optimization}

The SpectralMLP involves spectral color transformations and the SAUNet imitates integral operations for spectrum maps.
The objective function includes the following two items: 1) the weighted spectrum map reconstruction loss $\mathcal{L}_{\text{spectral}}$, which pushes the SpectralMLP to produce desired spectrum maps; 2) the RGB reconstruction loss $\mathcal{L}_{\text{RGB}}$, which optimizes the SAUNet to generate high-quality RGB images. The full objective function is described as:
\begin{equation}
\small
\mathcal{L}=\mathcal{L}_{\text{spectral}} + {\lambda}_{\text{RGB}} \mathcal{L}_{\text{RGB}},
\end{equation}
where ${\lambda}_{\text{RGB}}$ is a hyper-parameter to balance the contributions of the two losses. We empirically set it to 1.1.

\textbf{Weighted Spectrum Map Reconstruction Loss.} 
We found the image quality of generated spectrum maps is different, with its power concentrated in wavelength near 380$nm$ and 780$nm$ tends to be black, resulting in higher PSNR scores and closer $L_1$ distance. Therefore, we propose the weighted spectral reconstruction loss $\mathcal{L}_{\text{spectral}}$ to better acquire useful information from more informative spectrum maps distributed in the middle of visible light. Similar to NeRF, we simultaneously optimize a coarse model and a fine model, and the loss is
defined as:
\begin{equation}
\small
\label{eq:loss_spectral}
\begin{aligned}
\mathcal{L}_{\text{spectral}}= {\textstyle\sum_{i}^{s_{\text{num}}}} w_s \cdot {\textstyle\sum_{\mathbf{r} \in \mathcal{R}(\mathbf{P})}}  ( & ||\widehat{\mathbf{S}}_{\lambda_i}^{c}(\mathbf{r})-\mathbf{S}_{\lambda_i}(\mathbf{r})||_2^2 + \\ 
& ||\widehat{\mathbf{S}}_{\lambda_i}^{f}(\mathbf{r})-\mathbf{S}_{\lambda_i}(\mathbf{r})||_2^2),
\end{aligned}
\end{equation}
where 
$\mathbf{S}_{\lambda_i}$ represents the spectrum maps, $\mathcal{R}(\mathbf{P})$ is a set of camera rays at target position $\mathbf{P}$, $\widehat{\mathbf{S}}^{c}$ and $\widehat{\mathbf{S}}^{f}$ represent the spectrum maps generated in the coarse stage and fine stage, and
$w_s$ are the weights that are correlated to the PSNR scores of spectrum maps: 
\begin{equation}
\small
w_s = 2^{{P_\text{max}}/{P_{\lambda}}},
\end{equation}
where $P_\text{max}$ is the maximum PSNR value of $s_{\text{num}}$ spectrum maps, and ${P_{\lambda}}$ is the PSNR value of wavelength $\lambda$. 

\begin{table*}[!ht]
\centering
\resizebox{1.0\textwidth}{!}{
\begin{tabular}{c|cccc|cccc|cccc|cccc}
\toprule
      & \multicolumn{4}{c}{Spaceship}     & \multicolumn{4}{c}{Rover car}     & \multicolumn{4}{c}{Digger}          & \multicolumn{4}{c}{Piano} \\
      & PSNR$\uparrow$   & SSIM$\uparrow$   & LPIPS$\downarrow$  & $L_1$ $\downarrow$    & PSNR$\uparrow$   & SSIM$\uparrow$   & LPIPS$\downarrow$  & $L_1$ $\downarrow$    & PSNR$\uparrow$  & SSIM$\uparrow$  & LPIPS$\downarrow$  & $L_1$ $\downarrow$   & PSNR$\uparrow$ & SSIM$\uparrow$ & LPIPS$\downarrow$ & $L_1$ $\downarrow$ \\
\midrule
NeRF   & 30.126 & 0.9358 & \cellcolor{lightyellow}0.0275 & 3.0156 & 27.350 & 0.8920 & 0.0618 & 5.6619 & 30.658 & 0.9187 & 0.0413 & 2.9206 & 31.667 & 0.9239 & \cellcolor{lightorange}0.0394 & 3.4253 \\
Mip-NeRF & \cellcolor{lightyellow}31.495 & \cellcolor{lightyellow}0.9475 & 0.0535 & \cellcolor{lightyellow}2.5548 & \cellcolor{lightyellow}30.028 & \cellcolor{lightyellow}0.9210 & \cellcolor{lightyellow}0.0376 & \cellcolor{lightorange}2.7658  & \cellcolor{lightyellow}33.301 & \cellcolor{lightyellow}0.9290 & 0.0435 & \cellcolor{lightyellow}2.7835 &  31.872  & \cellcolor{lightorange}0.9304 & 0.0630 & \cellcolor{lightorange}2.8079\\
Aug-NeRF & 30.929 & 0.9402 & 0.0389 & 2.8378 & 27.275  & 0.9022 & 0.0512 & 5.0537  & 31.538 &  0.9248    & \cellcolor{lightyellow}0.0341  &  2.9024      &  31.876 & 0.9229   & 0.0471  &  3.1961  \\
Ours   & \cellcolor{lightorange}31.951 & \cellcolor{lightorange}0.9482 & \cellcolor{lightorange}0.0250 & \cellcolor{lightorange}2.5015 & \cellcolor{lightorange}30.086 & \cellcolor{lightorange}0.9212 & \cellcolor{lightorange}0.0356 & \cellcolor{lightyellow}3.2212 & \cellcolor{lightorange}33.378 & \cellcolor{lightorange}0.9357 & \cellcolor{lightorange}0.0259 & \cellcolor{lightorange}2.2897 & \cellcolor{lightorange}32.266 & \cellcolor{lightyellow}0.9290 & \cellcolor{lightyellow}0.0411 & \cellcolor{lightyellow}3.1049 \\
\midrule
\midrule
       & \multicolumn{4}{c}{Vintage car}     & \multicolumn{4}{c|}{Cartoon knight}     & \multicolumn{4}{c}{Kitchen}          & \multicolumn{4}{c}{Living room} \\
       & PSNR$\uparrow$   & SSIM$\uparrow$   & LPIPS$\downarrow$  & $L_1$ $\downarrow$    & PSNR$\uparrow$   & SSIM$\uparrow$   & LPIPS$\downarrow$  & $L_1$ $\downarrow$    & PSNR$\uparrow$  & SSIM$\uparrow$  & LPIPS$\downarrow$  & $L_1$ $\downarrow$   & PSNR$\uparrow$ & SSIM$\uparrow$ & LPIPS$\downarrow$ & $L_1$ $\downarrow$ \\
\midrule
NeRF   & 33.478 & 0.7958 & \cellcolor{lightorange}0.1319 & 3.2779 & 34.485 & 0.9273 & 0.1545 & 2.6076 & \cellcolor{lightyellow}34.583 & 0.8943 & 0.1650  & \cellcolor{lightorange}3.0413 & \cellcolor{lightyellow}33.172  & \cellcolor{lightyellow}0.9929  & \cellcolor{lightyellow}0.0578 & \cellcolor{lightyellow}3.2183 \\
Mip-NeRF & \cellcolor{lightyellow}33.883 & \cellcolor{lightyellow}0.8166 & 0.1747  & \cellcolor{lightorange}2.8633 & \cellcolor{lightorange}35.102  & 0.9572  & \cellcolor{lightyellow}0.1526 &   2.5882     & -- & -- & -- & -- &  -- & -- & -- & -- \\
Aug-NeRF & 33.639 &  0.8002 & 0.1536 & 3.3522 & 33.908 & 0.9287 & 0.1705 & 2.5693 &  34.480 & \cellcolor{lightyellow}0.9026  &  \cellcolor{lightyellow}0.1603  &   3.4897      &   32.205   & 0.9649  &   0.0706    &  4.3757  \\
Ours   & \cellcolor{lightorange}34.480 & \cellcolor{lightorange}0.8169 & \cellcolor{lightyellow}0.1499 & \cellcolor{lightyellow}2.8984 & \cellcolor{lightyellow}34.915 & \cellcolor{lightorange}0.9573 & \cellcolor{lightorange}0.1510 & \cellcolor{lightorange}2.4681 & \cellcolor{lightorange}35.115 & \cellcolor{lightorange}0.9117 & \cellcolor{lightorange}0.1637 & \cellcolor{lightyellow}3.0569 & \cellcolor{lightorange}33.665 & \cellcolor{lightorange}0.9931 & \cellcolor{lightorange}0.0479 & \cellcolor{lightorange}3.1378 \\
\bottomrule
\end{tabular}
}
\caption{Quantitative comparisons with other NeRF-based methods in terms of PSNR, SSIM, LPIPS and $L_1$ distance on 8 synthetic datasets. 
Orange indicates the best performance and Yellow refers to the second best result.
}
\label{tab:quantitative1}
\end{table*}

\begin{table}[!ht]
\centering
\resizebox{1.0\linewidth}{!}{
\begin{tabular}{c|cccc|cccc}
\toprule
       & \multicolumn{4}{c}{Projector}     & \multicolumn{4}{c}{Dog doll}   \\
       & PSNR$\uparrow$   & SSIM$\uparrow$   & LPIPS$\downarrow$  & $L_1$ $\downarrow$    & PSNR$\uparrow$   & SSIM$\uparrow$   & LPIPS$\downarrow$  & $L_1$ $\downarrow$  \\
\midrule
NeRF   & 28.9670 & 0.9429 &\cellcolor{lightyellow} 0.0472 & 7.6394 & \cellcolor{lightyellow}22.5040 & 0.8740 & \cellcolor{lightyellow} 0.1319 & 17.4331  \\
Aug-NeRF & \cellcolor{lightyellow}30.0795 &\cellcolor{lightorange} 0.9573  & \cellcolor{lightorange}0.0354 & \cellcolor{lightyellow} 5.8298 & 21.2073 & \cellcolor{lightyellow}0.8915 & \cellcolor{lightorange}0.1168 & \cellcolor{lightyellow}15.7774 \\
Ours   & \cellcolor{lightorange}31.2535 & \cellcolor{lightyellow}0.9449 & 0.0605 & \cellcolor{lightorange}5.7583 & \cellcolor{lightorange}25.1257 & \cellcolor{lightorange} 0.8932 &  0.14501 & \cellcolor{lightorange} 11.1261 \\
\bottomrule
\end{tabular}
}
\caption{Quantitative comparisons on 2 real-world scenes. 
}
\label{tab:quantitative3}
\end{table}

\textbf{RGB Reconstruction Loss.}
The RGB reconstruction loss $\mathcal{L}_{\text{RGB}}$ 
minimizes the difference between the predicted RGB image $\widehat{C}$ and the rendered RGB ground truth $C$. The $\mathcal{L}_2$ distance is adopted as the loss function, written as:
\begin{equation}
\small
\mathcal{L}_{\text{RGB}}=||\widehat{C}-C||_2^2.
\end{equation}


\section{Datasets and Implementations}

\subsection{Datasets}

\begin{figure}[t]
   \centering
   \includegraphics[width=0.95\linewidth]{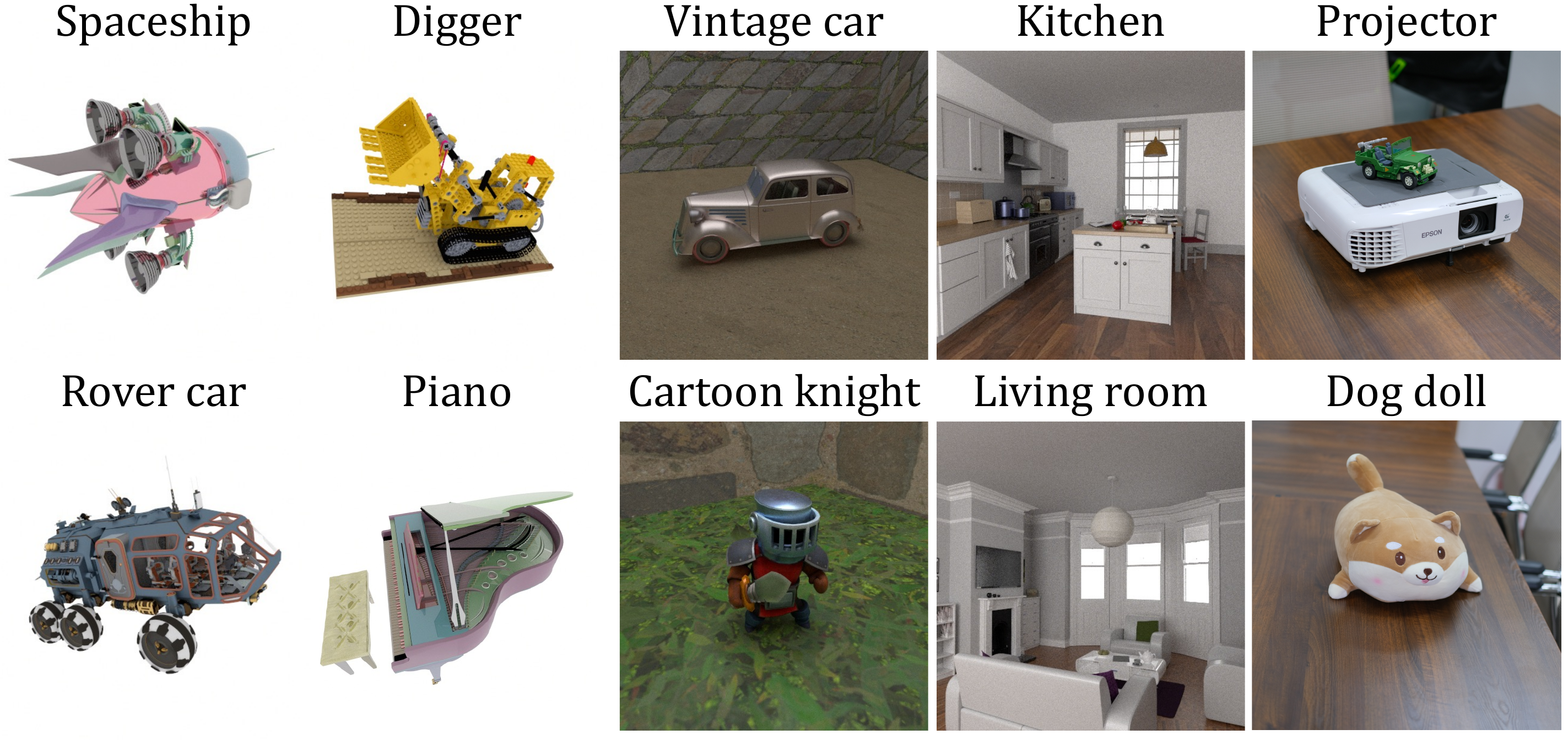}\\
   \caption{Scenes used for our synthetic and real datasets.
}
   \label{fig:dataset_examples}
\end{figure}

\textbf{Synthetic Scenes.} 
We first render 6 scenes with models located in the middle of the field, among which 4 scenes are designed without background and the viewpoints are captured on a sphere surrounding the models (the first two columns of Fig.~\ref{fig:dataset_examples}). 
The other 2 scenes use texture to construct the walls and floors, and their viewpoints are sampled on the upper hemisphere (the third column). 
Then, we render 2 indoor forward-facing scenes with limited camera location and 
camera perspectives (the fourth column). 

\textbf{Real-world Scenes.} We capture 2 forward-facing real datasets in a sealed room (the last column of Fig.~\ref{fig:dataset_examples}) using a camera and 8 color absorbers whose center wavelengths range from 400$nm$ to 750$nm$ with the interval of 50$nm$.
Different color absorbers are covered to the camera lens to obtain the spectral images.
Each real-world dataset includes approximately 40 viewpoints.

\subsection{Implementation Details}

We implement the SpectralMLP on top of NeRF~\cite{mildenhall2020nerf}, which uses an eight-layer MLP with 256 channels and ReLU activation to predict the density $\sigma$, and following two fully-connected layers with 128 and $3 \times s_{\text{num}}$ channels to obtain the spectral radiance. We sample 64 points along each ray in the coarse model and 128 points in the fine model on the dataset. Adam optimizer~\cite{kingma2014adam} is used for the SpectralMLP and the SAUNet, and their learning rate is set to $5 \times 10^{-4}$ and $0.001$, respectively.


\section{Experiments}

We conduct comprehensive experiments to verify the effectiveness of our method.
Specifically, we compare the SpectralNeRF against the following methods: 1) NeRF~\cite{mildenhall2020nerf}; 2) Mip-NeRF~\cite{barron2021mip}, an improved version of NeRF that reduces aliasing; 3) Aug-NeRF~\cite{chen2022aug}, which addresses the inherent non-smooth geometries of NeRF. 

\subsection{Quantitative Comparisons}\label{sec:exp_quantitative}

We report quantitative performance using PSNR (higher is better), SSIM (higher is better), LPIPS~\cite{zhang2018unreasonable} (lower is better), as well as $L_1$ distance (lower is better). 
Table~\ref{tab:quantitative1} shows the results 
on synthetic datasets, among which the top part is the results on 4 scenes without background, and the bottom part displays the results on 2 scenes using texture as background and 2 forward-facing scenes. Table~\ref{tab:quantitative3} shows the results on 2 real-world datasets. Note that, Mip-NeRF is not applicable to forward-facing cases. As for challenging real-world scenes, the SAUNet tends to produce slightly blurry results while preserving satisfactory image structure and contents, which leads to slightly lower LPIPS scores. This is a common trade-off in image restoration tasks.
Benefiting from the spectral radiance field, our method outperforms other 
methods because it simplifies complicated scenes. 
On average, our method achieves 5\% improvements compared to other NeRF-based methods.

\begin{figure*}[t]
   \centering
   \includegraphics[width=0.925\linewidth]{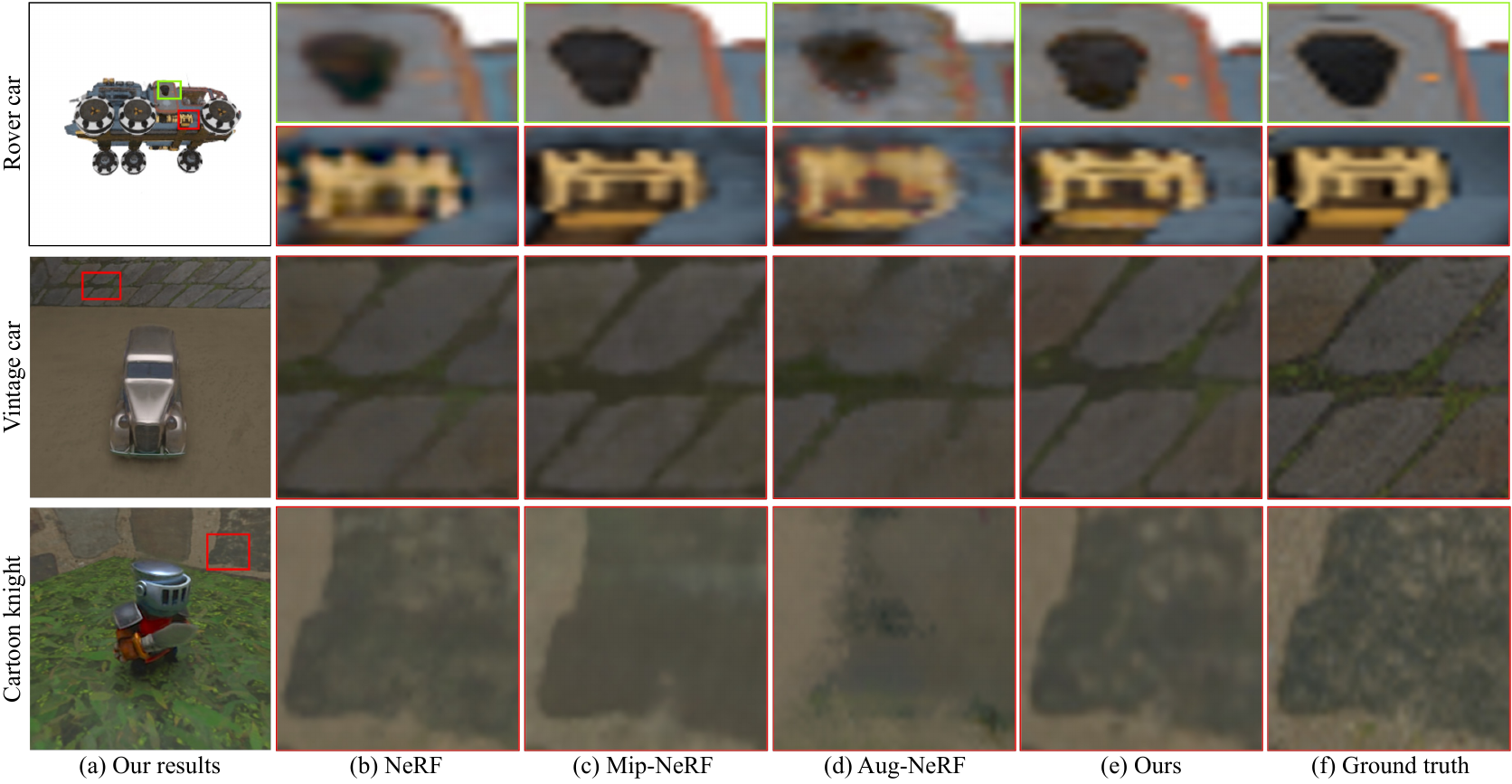}\\
   \caption{Qualitative comparisons with recent NeRF-based methods. In the first row, the comparison methods cannot recover the orange point, among which NeRF blurs it, Mip-NeRF and Aug-NeRF smooth it. 
   The SpectralNeRF
   preserves more details.
   }
   \label{fig:exp_qualitative}
\end{figure*}

\begin{figure}[t]
   \centering
   \includegraphics[width=1.0\linewidth]{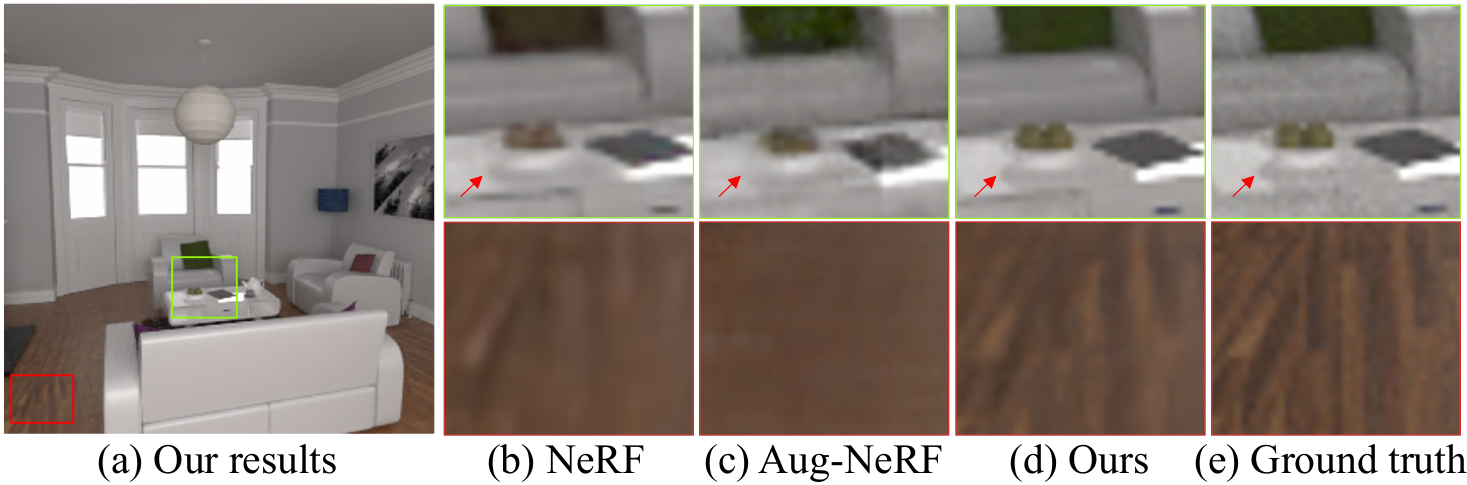}\\
   \caption{Comparisons on the synthetic forward-facing scene. NeRF and Aug-NeRF cannot reconstruct 
   the shading of the fruit platter and the texture of the floor.}
   \label{fig:exp_qualitative2}
\end{figure}

\begin{figure}[t]
   \centering
   \includegraphics[width=1.0\linewidth]{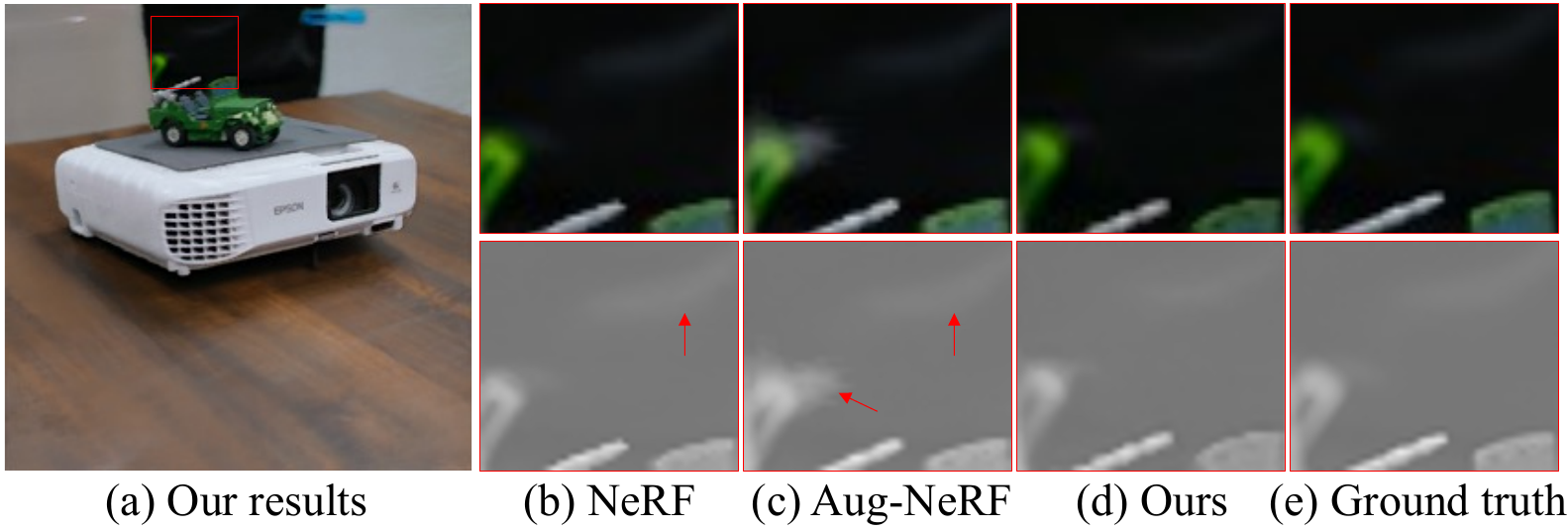}\\
   \caption{Comparisons on the real-world scene. NeRF and Aug-NeRF get results with fog artifacts. 
   We modify the contrast and the color to make the comparisons clear.}
   \label{fig:exp_qualitative3}
\end{figure}

\begin{figure}[!t]
   \centering
   \includegraphics[width=1.0\linewidth]{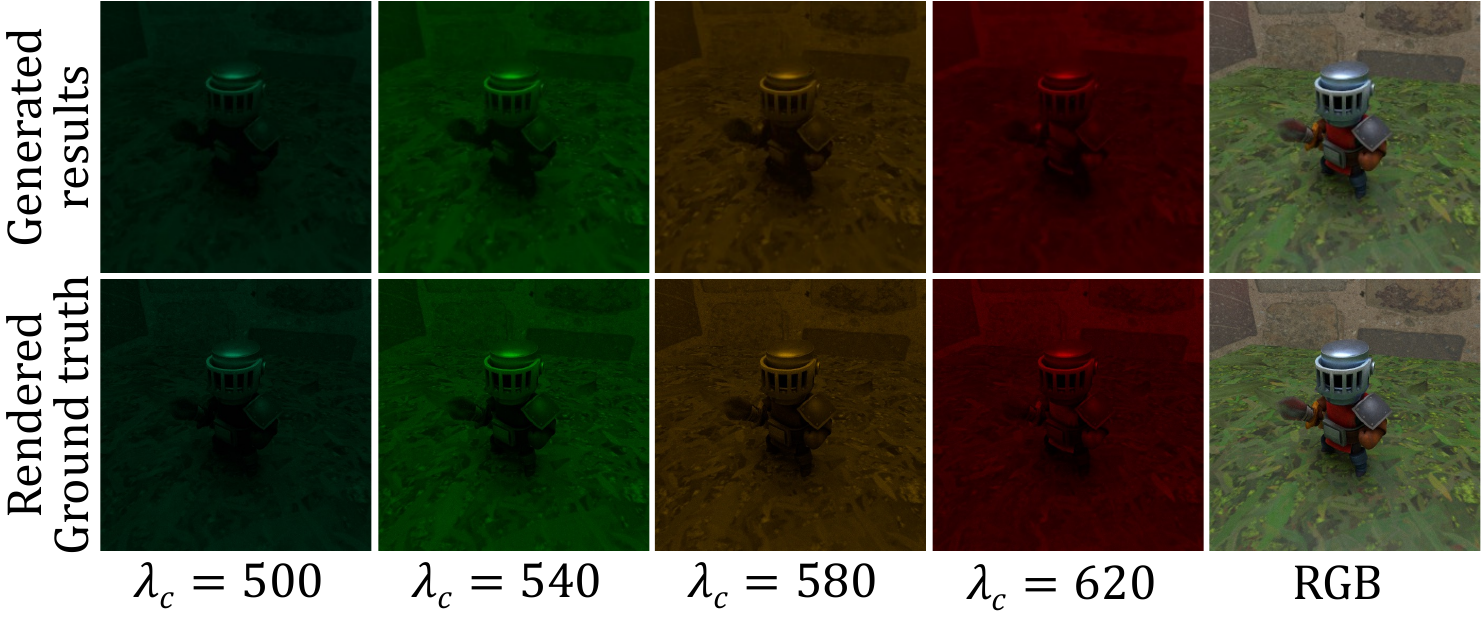}\\
   \caption{Several generated spectrum maps.
   }
   \label{fig:exp_snerf}
\end{figure}

\begin{table}[!ht]
\centering
\resizebox{1.0\linewidth}{!}{
\begin{tabular}{ccccc}
\toprule
            & NeRF  & Mip-NeRF & Aug-NeRF & Ours \\
\midrule
Time ($s$)        & 2.714 (-5\%)  & 14.054 (+393\%) & 8.366 (+193\%) & 2.852 (+0\%)                                                                   \\
\bottomrule
\end{tabular}
}
\caption{Inference time on testsets with resolution $256 \times 256$.}
\label{tab:exp_time}
\end{table}

\begin{table}[!ht]
\centering
\resizebox{0.95\linewidth}{!}{
\begin{tabular}{c|cccccccc}
\toprule
& $s_{\text{num}}$ & AG & SA & $w_s$ & PSNR$\uparrow$ & SSIM$\uparrow$ & LPIPS$\downarrow$  & $L_1$ $\downarrow$  \\
\midrule
(a) & 0               &                &         & \checkmark & 30.126        & 0.9358           & 0.0275             & 3.0156  \\
(b) & 11              &                &         & \checkmark & 31.375        & 0.9429           & 0.0285             & 2.7531 \\
(c) & 11              & \checkmark     &         & \checkmark  & 31.507        & 0.9445           & 0.0261             & 2.6085  \\
(d) & 11              & \checkmark    & E1         & \checkmark  & 31.552        & 0.9447           & 0.0266             & 2.6489   \\
(e) & 11              & \checkmark     & E1+E2+E3 & \checkmark  &  31.723      & 0.9446           & 0.0263             & 2.5892    \\
(f) & 11              & \checkmark     & E1+E2   &   & 31.601       & 0.9449           & 0.0268             & 2.6207  \\
(g) & 11              &      & E1+E2    & \checkmark  & \cellcolor{lightyellow}31.875        & \cellcolor{lightyellow}0.9479           & \cellcolor{lightorange}0.0244             & \cellcolor{lightyellow}2.5298  \\
(h) & 11              & \checkmark     & E1+E2   &  \checkmark  & \cellcolor{lightorange}31.951        & \cellcolor{lightorange}0.9482           & \cellcolor{lightyellow}0.0250             & \cellcolor{lightorange}2.5015   \\
\bottomrule
\end{tabular}
}
\caption{Ablation studies of different components. AG is the Attention Gate. SA is the Spectrum Attention module. E1, E2, and E3 are the first, second, and third encoder blocks. $w_s$ represents the weights in Eq.~\ref{eq:loss_spectral}.
}
\label{tab:exp_ablation}
\end{table}

\subsection{Qualitative Comparisons}\label{sec:exp_qualitative}

We present the qualitative comparisons of SpectralNeRF and 
NeRF-based methods in Fig.~\ref{fig:exp_qualitative}, Fig.~\ref{fig:exp_qualitative2} and Fig.~\ref{fig:exp_qualitative3}. NeRF may cause aliasing when rendering views of varying resolutions. Mip-NeRF extends NeRF to instead reason about volumetric frustums along a cone, but fails to recover the detailed geometry and appearance when handling difficult cases. Aug-NeRF uses worst-case perturbations to regularize the model. However, the operation lacks robustness, and sometimes brings negative effects. 
Results in Fig.~\ref{fig:exp_qualitative} clearly demonstrate that our method generates new viewpoints which are the closest to the ground truth. 
Other methods can reconstruct the low-frequency geometry, but fail to generate high-quality fine details. The results of NeRF and Aug-NeRF in Fig.~\ref{fig:exp_qualitative2} and Fig.~\ref{fig:exp_qualitative3} also cannot reconstruct fine details of the scene and generate results with fog artifacts. By concentrating on different spectral components of the complicated scene, our method outperforms the comparison methods by preserving more details. We then exhibit several generated spectrum maps of the SpectralMLP in Fig.~\ref{fig:exp_snerf}.

\subsection{Computational Times}\label{sec:exp_time}

The comparisons of inference times 
are reported in Table~\ref{tab:exp_time}. 
Mip-NeRF and Aug-NeRF design complicated technologies to improve their performance, and therefore costing more time. Some recent methods focus on efficient training or inference. Nevertheless, it is challenging to achieve superiority in both speed and quality. The speed of our method is comparable to general NeRF methods. Our method is a little slower than NeRF. The volume rendering for $s_\text{num}$ spectrum maps takes most of the extra time. 

\subsection{Ablation Studies}\label{sec:exp_ablation}

\textbf{Effectiveness of Main Components.}
We conduct ablations on several components to understand how these main modules work, including
$s_{\text{num}}$,
Attention Gate,
SA module,
and the weights $w_s$ in Eq.~\ref{eq:loss_spectral}. The results are shown in Table~\ref{tab:exp_ablation}. First, as shown in Table~\ref{tab:exp_ablation} (a) and (b), introducing the spectral radiance fields can effectively improve the performance. Second, as shown in Table~\ref{tab:exp_ablation} (g) and (h), removing the Attention Gate (AG) will degrade the results. Third, as shown in Table~\ref{tab:exp_ablation} (f) and (h), removing the weights $w_s$ in Eq.~\ref{eq:loss_spectral} also affects the performance. Fourth, Table~\ref{tab:exp_ablation} (c), (d), (e), and (h) show the results when embedding the SA module to different encoder blocks. The SA module is placed in the first two encoders to best explore the correlations of spectrum maps.

\begin{table}[t]
\centering
\resizebox{1.0\linewidth}{!}{
\begin{tabular}{c|cccc|c}
\toprule
       & Rover car & Digger  & Vintage car & Cartoon knight & Mean \\
\midrule
FC     & 27.309    & 30.617 & 33.382      & 34.413 &  31.430    \\
SAUNet & 30.086    & 33.378 & 34.480      &  34.915 & 33.214  \\
\bottomrule
\end{tabular}
}
\caption{The PSNR scores for the FC and the SAUNet on several datasets.}
\label{tab:aba_fc}
 
\end{table}

\begin{table}[!ht]
\centering
\resizebox{1.0\linewidth}{!}{
\begin{tabular}{c|cccc|c}
\toprule
       & Digger (PNG) & Vintage car & Kitchen  & Projector  & Mean \\
\midrule
Aug-NeRF     & 31.379   & 33.639  & 34.480  & 30.080  &  31.157    \\
S-Aug-NeRF & 31.506 & 34.130   & 34.974    &  32.385   & 32.636  \\
\midrule
Improvements  & +0.127 & +0.491 & +0.494 & +2.305    &  +0.854 \\
\bottomrule
\end{tabular}
}
\caption{The improvements when selecting Aug-NeRF as the baseline of our SpectralMLP.}
\label{tab:aba_improvements}
\end{table}

\textbf{The Role of SAUNet.} 
We conduct the ablation study when replacing the SAUNet network with a simple full-connect (FC) layer to demonstrate the importance and effectiveness of the SAUNet. Table~\ref{tab:aba_fc} shows corresponding results of several datasets, and the performance of the FC layer is obviously inferior to the SAUNet.

\textbf{Ablation Study of MLP.}
The SpectralNeRF can be considered as an improvement technique of existing 
methods, which can promote their performance if we apply their network architecture as the baseline of SpectralMLP. We conduct the experiments when transferring the spectral radiance fields and SAUNet to Aug-NeRF (named S-Aug-NeRF), and the improvements of PSNR scores are listed in Table~\ref{tab:aba_improvements}.

\begin{figure}[!ht]
   \centering
   \includegraphics[width=1.0\linewidth]{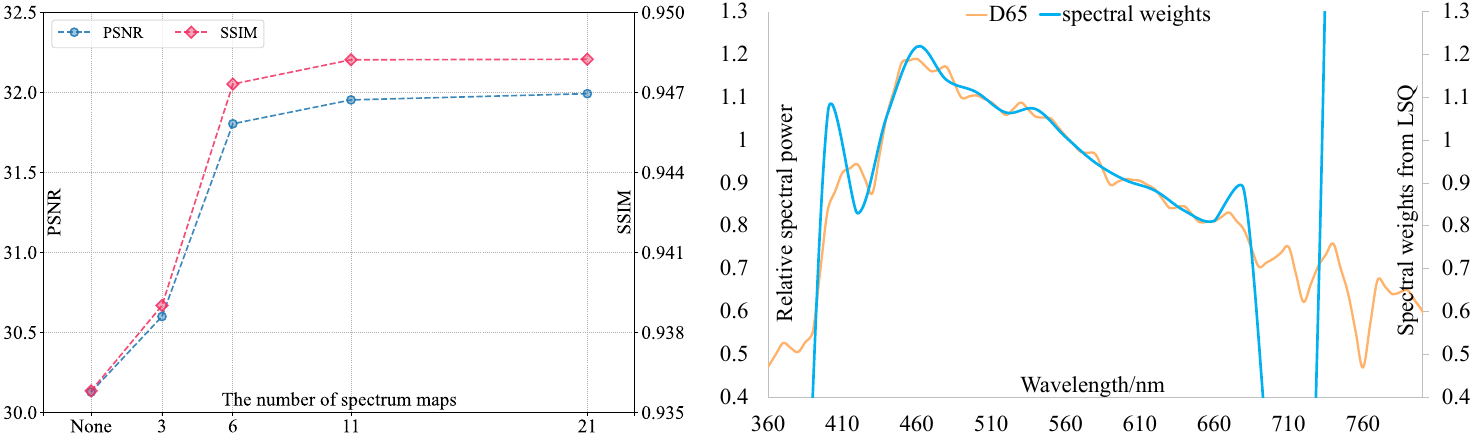}\\
   \caption{Left: The results when selecting different number of spectrum maps. Right: The correlations between CIE standard illuminant D65 and spectral weights.}
   \label{fig:spectrum_num_weight}
\end{figure}

\textbf{The Number of Spectral Maps.} 
Figure~\ref{fig:spectrum_num_weight} (Left) shows the PSNR and SSIM scores when setting the number of spectrum maps to 0 (vanilla NeRF), 6, 11, and 21. 
Obviously, reducing the output dimension (11 $\to$ 6) 
slightly affects the performance, while the effect of changing from 11 to 21 is minor. To achieve the balance,
we set the number of spectrum maps to 11 for the synthetic datasets and 8 for the real-world datasets. Second, Fig.~\ref{fig:spectrum_num_weight} (Left) shows the results when setting the outputs of the SpectralMLP as three RGB channels, where the SAUNet is considered as a refinement network. That is, the spectral loss is removed. The results (3-dimension) are superior to the vanilla NeRF, while inferior to results with spectral information.

\textbf{The Variant Spectral Rendering Pipeline.}
To verify the correctness of the variant spectral rendering
pipeline, we apply the least square method to obtain the adapted weights for spectrum maps of wavelength bands.The estimated weights should be approximately proportional to the spectral power distribution $L(\lambda)$ of CIE standard illuminant D65 except for bands near 380$nm$ and 780$nm$ which are less informative. The almost identical curves in Fig.~\ref{fig:spectrum_num_weight} (Right) demonstrate the rationality of the proposed variant spectral rendering. 


\section{Conclusion}

We have proposed SpectralNeRF, an end-to-end NeRF-based method to achieve 
physically-based spectral rendering. We modified the traditional spectral rendering pipeline into two steps and designed SpectralMLP and SAUNet to build up the two steps. With the help of the spectral radiance field, our method can generate high-quality RGB output of white-light illumination. Comprehensive experiments have demonstrated the superiority of the proposed SpectralNeRF.
In the future, we plan to reduce the number of spectrum maps, and construct each view with only one or sparse discrete spectral maps, which will broaden the applications.


\end{document}